\documentclass{article}
\usepackage{spconf,amsmath,graphicx, amsfonts}
\usepackage{algorithm}
\usepackage{algpseudocode}
\usepackage{enumitem}
\usepackage{hyperref, lmodern}
\usepackage{url}
\usepackage{mwe} 
\usepackage{multirow}
\usepackage{titlesec}

\title{Tell2Reg: Establishing spatial correspondence between images by the same language prompts}

\name{
\parbox{\linewidth}{\centering
Wen Yan$^{1,2}$, Qianye Yang$^{1,2}$, Shiqi Huang$^{1,2}$, Yipei Wang$^{1,2}$, Shonit Punwani$^{5}$, Mark Emberton$^{6}$, Vasilis Stavrinides$^{3,4,7}$, Yipeng Hu$^{1,2}$, Dean Barratt$^{1,2}$}}

\address{
\parbox{\linewidth}{\centering
\{$^{1}$UCL Hawkes Institute; 
$^2$Department of Medical Physics and Biomedical Engineering;
$^3$Cancer Institute;
$^4$Urology Department, UCL Hospital;
$^{5}$Centre for Medical Imaging, Division of Medicine;
$^{6}$Division of Surgery and Interventional Science\}, University College London\\
$^7$ Radiology Department, Imperial College Healthcare
}}

\begin{document}
%
\maketitle
\begin{abstract}
Spatial correspondence can be represented by pairs of segmented regions, such that the image registration networks aim to segment corresponding regions rather than predicting displacement fields or transformation parameters. In this work, we show that such a corresponding region pair can be predicted by the same language prompt on two different images using the pre-trained large multimodal models based on GroundingDINO and SAM. This enables a fully automated and training-free registration algorithm, potentially generalisable to a wide range of image registration tasks. In this paper, we present experimental results using one of the challenging tasks, registering inter-subject prostate MR images, which involves both highly variable intensity and morphology between patients. Tell2Reg is training-free, eliminating the need for costly and time-consuming data curation and labelling that was previously required for this registration task. This approach outperforms unsupervised learning-based registration methods tested, and has a performance comparable to weakly-supervised methods. Additional qualitative results are also presented to suggest that, for the first time, there is a potential correlation between language semantics and spatial correspondence, including the spatial invariance in language-prompted regions and the difference in language prompts between the obtained local and global correspondences. Code is available at \url{https://github.com/yanwenCi/Tell2Reg.git}.

\end{abstract}
\section{Introduction}
\label{sec:intro}
\begin{figure}[htb]
    \centering
    \includegraphics[width=0.7\linewidth]{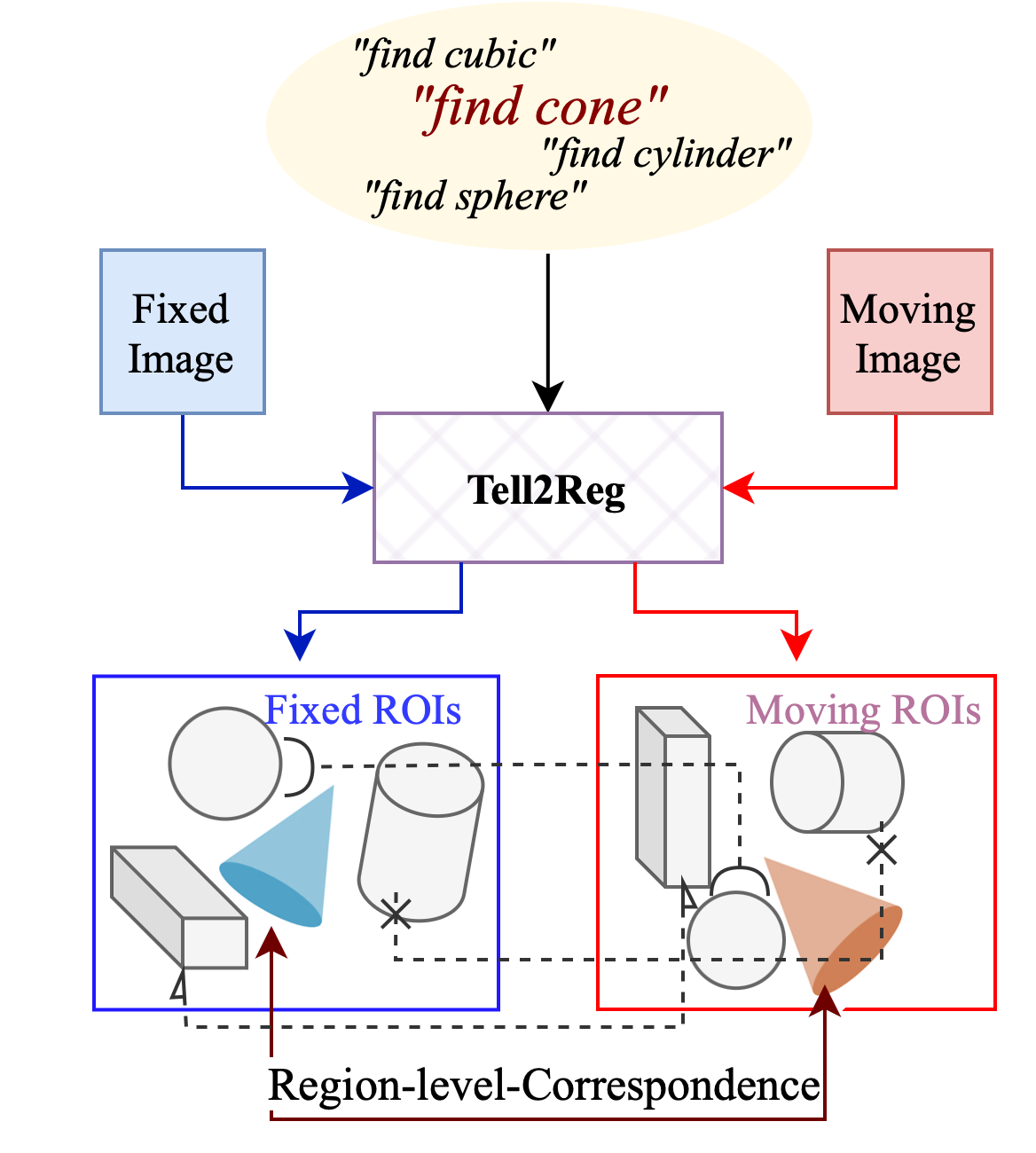}
    \caption{A brief illustration of how Tell2Reg framework uses text prompts to generate corresponding ROIs from fixed and moving images.}
    \label{fig:abstract}
\end{figure}
 
MR imaging has increasingly been used for diagnosing prostate cancer, planning targeted biopsy and other treatment procedures. Registering images from different patients is an interesting research topic, which may enable propagating procedural plans from reference images to new patients, constructing MR-based lower-pelvic atlases and other similar population studies~\cite{harris2015consensus}. Among the sequences that are useful for prostate cancer diagnosis, T2-weighted sequences contain the richest soft tissue contrast and are often used in these registration tasks, which is a focus of the previous work as well as this study \footnote{The potential need, albeit comparably minor, of inter-sequence registration has been a research topic of its own~\cite{xu2016evaluation} and is not discussed further in this paper.}. Registering inter-subject lower-pelvic MR images is challenging due to significant variability in intensity and morphology between subjects, unlike the intra-subject registration tasks~\cite{yang2021morphological}. This task usually requires large training datasets and benefits from segmentation labels and advanced training strategies~\cite{9925717}. Classical iterative algorithms has been shown inferior to learning-based approaches in this application~\cite{9925717}.
Existing image registration methods often use deep learning models to learn pixel or feature correspondences between moving (to be aligned) and fixed (reference) images~\cite{balakrishnan2019voxelmorph,evan2022keymorph,chen2022transmorph}. These models typically optimize a similarity metric that measures how well the warped moving image aligns with the fixed image, based on intensity differences, such as cross correlation and mutual information. 
Recent work also incorporated large language models for the same optimisation goal~\cite{ma2024large, chen2024spatially}, to predict a deformation field or a parametric transformation. 
Following the discussion in a previous work~\cite{huang2024one} that connects region-level correspondence and the ability to segment a wide range of objects, with or without anatomical significance, this paper describes a novel approach to establish spatial correspondence by multimodal approaches with language prompts.
Notably, segment anything model (SAM)~\cite{kirillov2023segment} has 
been developed together with language models, such as CLIP~\cite{hafner2021clip} and Bert~\cite{kenton2019bert}, to generate regions of interest (ROIs) from text prompts. GroundingDINO~\cite{liu2023grounding} models have merged the multi-level language model with object detection, generating ROIs from text prompts via contrastive learning.  

We hypothesize that the image registration task does not need to optimise pixel-level correspondence during model training; instead, it can be conceptualized as detecting the same regions from different images, therefore leveraging capabilities of segmentation models such as SAM without retraining. Further, by utilising pre-trained multimodal systems, these corresponding regions ought to be directly predicted by prompting identical text in the proposed ``Tell2Reg'' approach, illustrated in Fig.\ref{fig:abstract}. The proposed image registration does not require any training or finetuning, outperforming the tested state-of-the-art learning-based registration methods.

\section{Method}
\label{sec:method}
\begin{figure}[htb]
\centering
\includegraphics[width=0.88\linewidth]{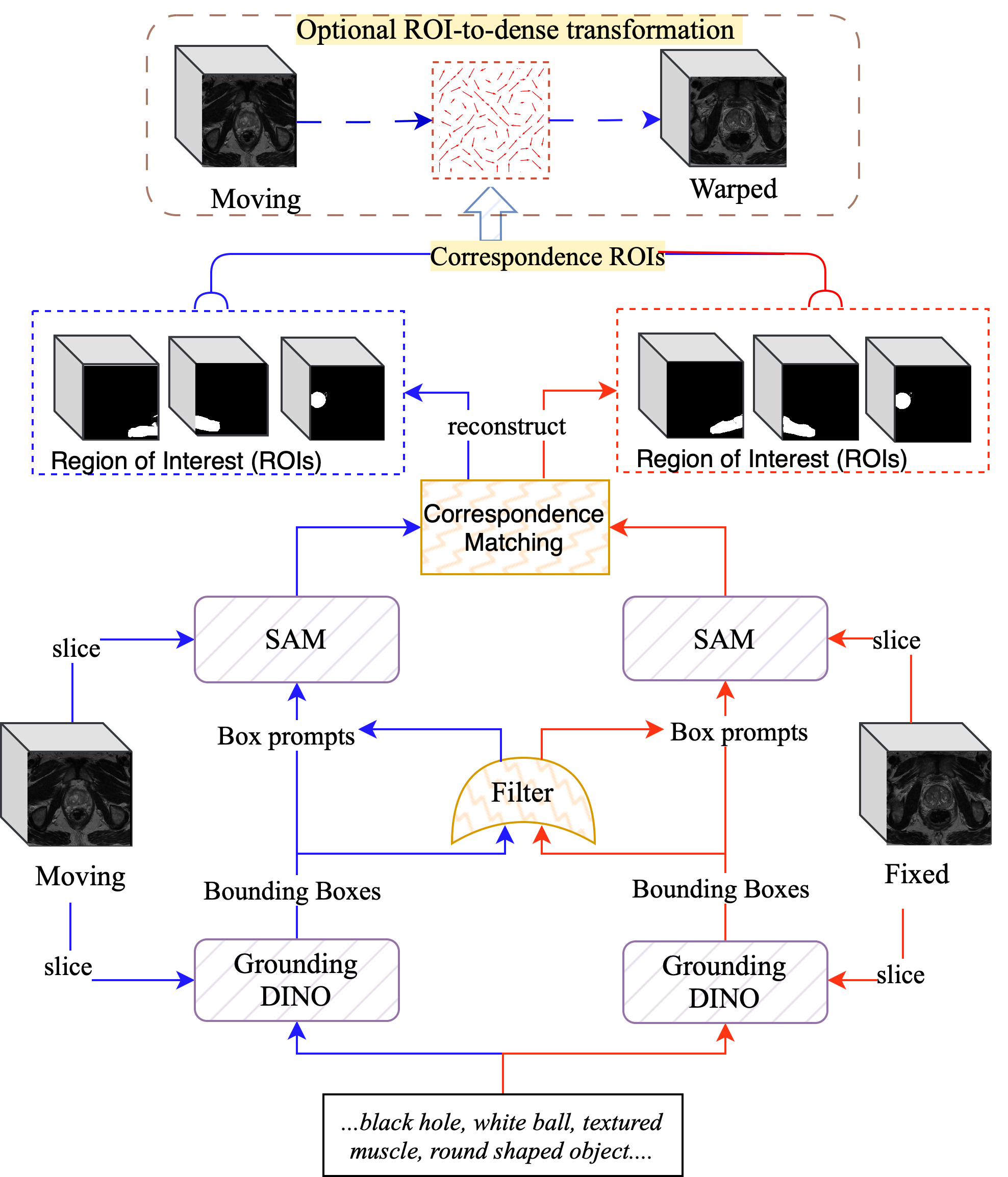}
\caption{Tell2Reg framework using pre-trained GroundingDINO and SAM. The ROI-to-dense transformation is an optional step for comparing with other methods.}
\label{fig:net}
\end{figure}

The above assertion - segmenting regions on both fixed and moving images using identical text prompts yields corresponding ROIs - represents an ideal scenario rather than a guaranteed outcome. 
With practical constraints such as the availability of medical-image-finetuned foundation models, selecting ROIs may require further prompt engineering and/or post-processing to ensure that the segmented ROIs are corresponding pairs. 
This section describes a specific algorithm, utilising pre-trained SAM and GroundingDINO, outlined in Fig.~\ref{fig:net}.

\subsection{Text to corresponding ROIs}
\label{sec:roi}
We start our discussion with a theorem \cite{huang2024one} stating that the registration task can be framed as a correspondence learning problem between regions. Let the fixed and moving images be  $I^{\textit{fix}}$ and  $I^{\textit{mov}}$, respectively. Establishing spatial correspondence between the two images can be achieved by identifying \( K \) pairs of corresponding ROIs, denoted as \( \{(R_k^{\textit{fix}}, R_k^{\textit{mov}})\}_{k=1}^K \). In the limiting case where each ROI reduces to a single pixel, the ROI-based correspondence becomes equivalent to a pixel-wise one, as governed by a dense displacement field (DDF). 

We use GroundingDINO~\cite{liu2023grounding} to generate bounding boxes ${B}^{\textit{fix}}=f_{dino}(I^{\textit{fix}}, p)$ and ${B}^{\textit{mov}}=f_{dino}(I^{\textit{mov}}, p)$ for fixed and moving images with same text prompt $p$, respectively, where $p$ can be any text, examples are given in Table~\ref{tab:prompts}.
Ideally, bounding boxes generated from identical text prompts should correspond precisely between the fixed and moving images; however, due to the limits of the GroundingDINO model on prostate data,  which could lead to false positive correspondence, by detecting regions that appear similar based on texture or intensity but are not corresponding regions of registration interest.

Before inputting BBox $B=\{B^{fix}, B^{mov}\}$ into the SAM model, we filter out those that are too small or too large, for reducing computation costs and potential outliers.
We then use the remaining bounding boxes as box prompts to segment ROIs in both fixed and moving images, to obtain the segmented ROIs for fixed and moving: \({R}^{\textit{fix}}=f_{sam}(I^{\textit{fix}}, {B}^{\textit{fix}})\) and \({R}^{\textit{mov}}=f_{sam}(I^{\textit{mov}}, {B}^{\textit{mov}})\)  
resulting in $\{M^{\textit{fix}}\}_{k=1}^{K^{\textit{fix}}}$ and $\{M^{\textit{mov}}\}_{k=1}^{K^{\textit{mov}}}$ as the binary masks of the ROIs in the fixed and moving images, respectively.


\subsection{ROI correspondence refinement}
\label{ssec:reg}
To establish robust correspondence between these ROIs, we propose examining the similarity between ROI prototypes from the same or different text prompts. The ROI prototypes are generated by averaging the ROIs in the fixed and moving images. Given two sets of ROIs, \(\{{R}^{\textit{fix}}\}_{k=1}^{K^{\textit{fix}}}\) and \(\{{R}^{\textit{mov}}\}^{K^{\textit{mov}}}_{k=1}\), where $K^{\textit{mov}}$ may not be equal to $K^{\textit{fix}}$, the proposed matching process finds the correspondence ROIs by minimizing the $L^2$ distance between the ROI prototypes, detailed in Algorithm~\ref{alg:correspondence}, yielding $K^{\textit{cor}}$ pairs of matched 
ROIs $\{(R_k^{\textit{fix}}, R_k^{\textit{mov}})\}_{k=1}^{K^{\textit{cor}}}$. It is interesting to discuss that, when the specificity of the multimodal segmentation system (Sec.~\ref{sec:roi}) improves, this refining step may render itself unnecessary, while the differences to the previous work~\cite{huang2024one} include better-selected ROI sets (due to the correspondence-informing text prompts) and thus a different similarity function.
These ROIs $\{(R_k^{\textit{fix}}, R_k^{\textit{mov}})\}_{k=1}^{K^{\textit{cor}}}$ represent the region-level correspondence between the fixed and moving images.

\begin{algorithm}[!htb]
\label{alg:correspondence}
  \caption{ROI correspondence calculation}
  \begin{algorithmic}[1]
  \State \textbf{Input}: Image $I^{\textit{fix}}, I^{\textit{mov}}$, ROI masks $\{{M}_k^{\textit{fix}}\}_{k=1}^{K^{\textit{fix}}}$, $\{{M}_k^{\textit{mov}}\}_{k=1}^{K^{\textit{mov}}}$, Pretrained SAM model $f_{sam}$

  \State \textbf{Output}: Paired ROIs $\{R^{\textit{fix}}_k, R^{\textit{mov}}_k\}_{k=1}^{K^{\textit{cor}}}$ 
  
  \State \textbf{Step 1: Compute ROI embeddings}
  
  \For{each mask ${M}_i^{fix}$}
    \State Compute image embeddings ${E^{\textit{fix}}}$ by $f_{sam}(I^{\textit{fix}})$.
    \State Resize mask ${M}_i^{fix}$ to size of $E^{\textit{fix}}$ as $M_i^{'fix}$:
      \State Multiply the embedding by the mask to isolate ROI features:
      \(
      {E}_{roi}^{fix} = {E}^{fix} \cdot {M}_i^{'fix}
      \)
      \State Compute prototypes within ROI for both fix and moving:
      \(
      {e}_i^{\textit{fix}} = \frac{1}{|{M}_i'|} \sum_{x,y} {E}_{roi}^{fix}[x,y]
      \).
\EndFor
\For {each mask $M_j^{mov}$}
    \State Do 5-8 for moving embedding, yielding \({e}_j^{\textit{mov}}\).
    \EndFor
  
  \State \textbf{Step 2: Compute Similarity Matrix}
  
  \For{each pair of embeddings $({e}_i^{\textit{fix}}, {e}_j^{\textit{mov}})$}
      \State Compute cosine similarity:
      \(
      S[i, j] = \frac{{e}_i^{\textit{fix}} \cdot {e}_j^{\textit{mov}}}{\|{e}_i^{\textit{fix}}\| \|{e}_j^{\textit{mov}}\|}
      \)
  \EndFor
   
  \State \textbf{Step 3: Find best correspondence}
  \State Paired ROIs: $\{R^\textit{{fix}}_k, R^{\textit{mov}}_k\}_{k=1}^{K^{\textit{cor}}}
=\arg\max\limits_{{(i,j)}}(S[i,j])$
  
  \State \textbf{Return}: Paired ROIs $\{R^\textit{{fix}}_k, R^{\textit{mov}}_k\}_{k=1}^{K^{\textit{cor}}}$

  \end{algorithmic}
  \end{algorithm}

\subsection{Optional dense transformation}
\label{ssec:roi2dense}
When useful, the region-level correspondence (represented by the ROI pairs, Sec.~\ref{ssec:reg}) can be converted to voxel-level dense correspondence (represented by a DDF), by iteratively minimising a region-specific alignment error $\mathcal{L}_{roi}$ and a $L^2$ regularization term $\mathcal{L}_{reg}$:
$\mathcal{L} = \mathbb{E}_{k}[\mathcal{L}_{roi}({R}_k^{fix}, \mathcal{T}({R}_k^{mov}, \Theta))] + \lambda \mathcal{L}_{reg}(\Theta)$, 
where $\mathbb{E}_{k}[\cdot]$ is the mathematical expectation, $\Theta$ is the parameters of the transformation model (here, the DDF), and $\lambda$ is the regularization weight. The region-specific alignment error $\mathcal{L}_{roi}$ is an equally weighted Dice and MSE loss.

\section{Experiments and Results}
\label{sec:results}
\subsection{Dataset and implementation details}
\label{ssec:data}
The five hundred and forty-two pairs T2 pelvis mpMR images were acquired from 850 prostate cancer patients, part of several clinical trials conducted at University College London Hospital. The images were resampled to 1mm isotropic resolution, sized $200 \times 200 \times 96$. Prostate segmentation masks labelled by experts are readily available for evaluation and ablation studies. The data were split to training, validation and test sets by 365, 90 and 87 pairs respectively.
 As our proposed method is totally training-free, we only use 87 pairs test data, while the other methods use all datasets. The proposed model was implemented with Pytorch, using pre-trained models~\cite{liu2023grounding,kirillov2023segment}, described in Sec.~\ref{sec:roi}.  Dice and target registration error (TRE) of centroids are used to evaluate the registration performance. The detection ratio describes the proportion of prostate ROI correspondences detected using a specific text prompt: ${N^{ROI^{cor}}_{prostate}}/{N_{prostate}}$, where $N^{ROI^{cor}}_{prostate}$ is detected corresponding prostate ROIs, and $N_{prostate}$ is the total number of prostate in ground-truth. 
\begin{figure*}[!htb]
\centering  
\includegraphics[width=0.88\linewidth]{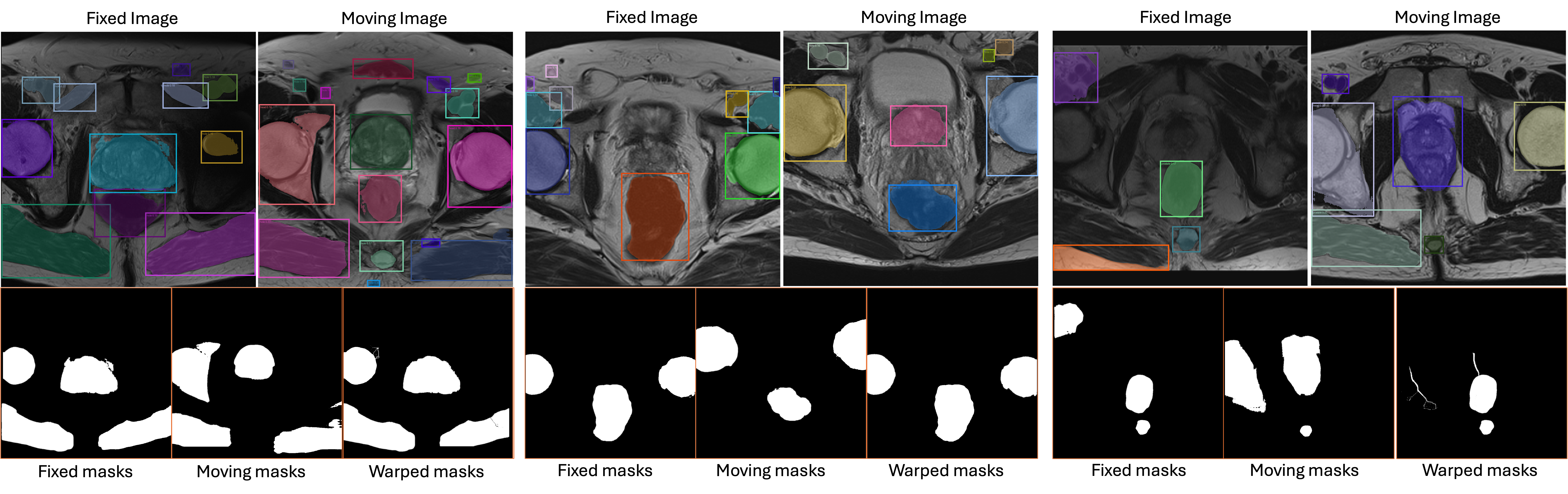}
\caption{Visualization of registration results. The bounding boxes were generated by GroundingDINO, and the coloured masks were produced by SAM. The binary masks highlight the selected corresponding regions of interest (ROIs) in the fixed and moving images alongside the warped masks to show the optional spatial alignment results. 
The first two groups show corresponded ROIs, and the third group shows bad example of mismatched ROIs.
}\label{fig:visual}
\end{figure*}
\subsection{Comparison experiments}
We compare the proposed method with the state-of-the-art registration methods, including VoxelMorph~\cite{balakrishnan2019voxelmorph}, KeyMorph~\cite{evan2022keymorph} and TransMorph~\cite{chen2022transmorph}, listed in Table~\ref{tab:results}. All the compared methods also required substantial training data and computation costs for training. The results showed that Tell2Reg outperforms unsupervised methods. As the only exception, weakly supervised TransMorph, which requires prostate segmentation labels, led to a higher average Dice without statistical significance, than that from Tell2Reg ($p=0.060$). It is also arguable that the TREs may be more indicative of registration performance, in many applications. 
Fig.~\ref{fig:visual} shows the visualization of the proposed Tell2Reg outputs. Prior work~\cite{li2022few} showed that an atlas from similar inter-subject registration improved pathology-indicating statistics (p=0.009). The significant improvement reported here highlights strong clinical potential to explore.

\begin{table}[!htb]
\caption{Comparison between the proposed and SOTA registration methods, which employs unsupervised or weakly supervised training using prostate masks, whereas our method is training-free.}
\label{tab:results}
\resizebox{\linewidth}{!}{
\begin{tabular}{c|cc|cc}
\hline
 & \multicolumn{2}{c|}{Unsupervised} & \multicolumn{2}{c}{Weakly Supervised} \\
Methods & Dice & TRE(mm) &  Dice & TRE(mm)\\
\hline
Before register & 0.68$\pm$0.21 & 11.12$\pm$2.22 & / & / \\
VoxelMorph & 0.69$\pm$0.11 & 8.06$\pm$1.67 & 0.78$\pm$0.08 &4.23$\pm$3.19 \\
KeyMorph &  0.73$\pm$0.09 & 5.96$\pm$2.82 &0.84$\pm$0.04 & 3.32$\pm$ 1.88 \\
TransMorph & 0.68$\pm$0.10 & 7.24$\pm$3.72 &0.85$\pm$0.04 & 4.25$\pm$1.90   \\
Tell2Reg &0.84$\pm0.06$ & 3.09$\pm$2.25&/&/ \\ \hline
\end{tabular}}
\end{table}

\begin{table}[!bt]
    \caption{Effect of text prompts on prostate ROI detection and correspondence. The 1st row shows the \textit{detection ratio} of prostate ROIs with different prompts. Remaining rows report the number of ROIs detected in the fixed and moving images and the count of corresponding ROIs between the two images.}
    \label{tab:prompts}
    \centering
    \resizebox{\linewidth}{!}{
    \begin{tabular}{ccccccc}
    \hline
        Detected ROIs        &\textit{``hole''}&\textit
{``head''}  & \textit{``prostate''} & \text{``dog''} & \textit{``correspond''} & \textit{``middle''}\\\hline
         Prostate ROI& 35\% &11\% &25\% & 19\% & 25\% & 23\%\\
         \hline
         ROIs Moving& 3279 & 1094&  1128 & 600& 427 & 452\\
         ROIs Fixed & 3651 & 1300& 1184& 678& 464 & 553\\
         Corresponding &1700&706& 715 &342 & 231 &271\\ \hline
    \end{tabular}}
\end{table}


\subsection{Text prompts comparison}

We first test the registration performance by changing the text prompts and how they are sampled. We summarise our experience as follows:
First, for a specific registration task, the performance is more stable with a fixed set of prompts, predefined empirically and qualitatively with a small set of independent data, compared with those based on randomly sampled (individually or in-pairs) text prompts. However, we are open to explore automatic text prompt refinement in future; 
Second, descriptive phrases generally lead to less corresponding ROIs than concrete or abstract nouns do, as those used in this study. 
Third, perhaps as expected, words and concepts related more closely to natural images (the models were trained with), such as \textit{``cat"}, \textit{``dog"}, \textit{``car"} and \textit{``tree"}, produced fewer sensible ROIs, compared with those found in medical images or more abstract, such as \textit{``hole"},  \textit{``prostate"} and \textit{``middle"}. However, how much each category contributed to useful correspondence remains unclear and an interesting open research question.
Table~\ref{tab:prompts} summarises a list of text prompts used to produce quantitative results reported in this paper, together with ratios between individual text-prompted ROIs overlapping with the prostate gland, as a reference example. The quantities of prompted ROIs and corresponding pairs after refinement are also summarised in the same table. 


\section{Discussion and Conclusion}
\label{sec:conclusion}
In this paper, we propose a novel registration method which learns the correspondence between regions, by utilising the existing SAM-based object detection model with text prompts. This approach does not require retraining, and shows better generalisability, compared with methods that even require substantial training and training data. Our future work will focus on exploring prostate-specific multimodel foundation models, developing automated prompt engineering, and implementing refinement strategies to enhance the robustness, adaptability, and efficiency of the proposed method. 

\section*{Acknowledgments}
\label{sec:acknowledgments}

This work is supported by the International Alliance for Cancer Early Detection, an alliance between Cancer Research UK [C28070/A30912; C73666/A31378; EDDAMC-2021/100011], Canary Center at Stanford University, the University of Cambridge, OHSU Knight Cancer Institute, University College London and the University of Manchester.

\section*{Compliance with Ethical Standards}
All trial patients gave written consents, with ethics approved as part of the respective trial protocols~\cite{hamid_smarttarget}.
\bibliographystyle{IEEEbib}
\bibliography{strings,refs}

\begin{thebibliography}{10}

\bibitem{harris2015consensus}
Victoria~A Harris, John Staffurth, Olivia Naismith, Alikhan Esmail, et~al.,
\newblock ``Consensus guidelines and contouring atlas for pelvic node delineation in prostate and pelvic node intensity modulated radiation therapy,''
\newblock {\em International Journal of Radiation Oncology* Biology* Physics}, vol. 92, no. 4, pp. 874--883, 2015.

\bibitem{xu2016evaluation}
Zhoubing Xu, Christopher~P Lee, Mattias~P Heinrich, et~al.,
\newblock ``Evaluation of six registration methods for the human abdomen on clinically acquired ct,''
\newblock {\em IEEE Transactions on Biomedical Engineering}, vol. 63, no. 8, pp. 1563--1572, 2016.

\bibitem{yang2021morphological}
Qianye Yang, Tom Vercauteren, Yunguan Fu, Giganti, et~al.,
\newblock ``Morphological change forecasting for prostate glands using feature-based registration and kernel density extrapolation,''
\newblock in {\em 2021 IEEE 18th International Symposium on Biomedical Imaging (ISBI)}. IEEE, 2021, pp. 1072--1076.

\bibitem{9925717}
Alessa Hering, Lasse Hansen, Tony~CW Mok, Albert~CS Chung, Siebert, et~al.,
\newblock ``Learn2reg: Comprehensive multi-task medical image registration challenge, dataset and evaluation in the era of deep learning,''
\newblock {\em IEEE Transactions on Medical Imaging}, vol. 42, no. 3, pp. 697--712, 2023.

\bibitem{balakrishnan2019voxelmorph}
Guha Balakrishnan, Amy Zhao, Mert~R Sabuncu, John Guttag, and Adrian~V Dalca,
\newblock ``Voxelmorph: a learning framework for deformable medical image registration,''
\newblock {\em IEEE transactions on medical imaging}, vol. 38, no. 8, pp. 1788--1800, 2019.

\bibitem{evan2022keymorph}
M~Yu Evan, Alan~Q Wang, Adrian~V Dalca, and Mert~R Sabuncu,
\newblock ``Keymorph: Robust multi-modal affine registration via unsupervised keypoint detection,''
\newblock in {\em Medical imaging with deep learning}, 2022.

\bibitem{chen2022transmorph}
Junyu Chen, Eric~C Frey, Yufan He, et~al.,
\newblock ``Transmorph: Transformer for unsupervised medical image registration,''
\newblock {\em Medical image analysis}, vol. 82, pp. 102615, 2022.

\bibitem{ma2024large}
Mingrui Ma, Weijie Wang, Jie Ning, Jianfeng He, Nicu Sebe, and Bruno Lepri,
\newblock ``Large language models for multimodal deformable image registration,''
\newblock {\em arXiv preprint arXiv:2408.10703}, 2024.

\bibitem{chen2024spatially}
Xiang Chen, Min Liu, Rongguang Wang, Renjiu Hu, et~al.,
\newblock ``Spatially covariant image registration with text prompts,''
\newblock {\em IEEE Transactions on Neural Networks and Learning Systems}, 2024.

\bibitem{huang2024one}
Shiqi Huang, Tingfa Xu, Ziyi Shen, Shaheer~Ullah Saeed, et~al.,
\newblock ``One registration is worth two segmentations,''
\newblock in {\em International Conference on Medical Image Computing and Computer-Assisted Intervention}. Springer, 2024, pp. 665--675.

\bibitem{kirillov2023segment}
Alexander Kirillov, Eric Mintun, Nikhila Ravi, Hanzi Mao, et~al.,
\newblock ``Segment anything,''
\newblock in {\em Proceedings of the IEEE/CVF International Conference on Computer Vision}, 2023, pp. 4015--4026.

\bibitem{hafner2021clip}
Markus Hafner, Maria Katsantoni, Tino K{\"o}ster, James Marks, et~al.,
\newblock ``Clip and complementary methods,''
\newblock {\em Nature Reviews Methods Primers}, vol. 1, no. 1, pp. 1--23, 2021.

\bibitem{kenton2019bert}
Jacob Devlin Ming-Wei~Chang Kenton and Lee~Kristina Toutanova,
\newblock ``Bert: Pre-training of deep bidirectional transformers for language understanding,''
\newblock in {\em Proceedings of naacL-HLT}. Minneapolis, Minnesota, 2019, vol.~1, p.~2.

\bibitem{liu2023grounding}
Shilong Liu, Zhaoyang Zeng, Tianhe Ren, et~al.,
\newblock ``Grounding dino: Marrying dino with grounded pre-training for open-set object detection,''
\newblock {\em arXiv preprint arXiv:2303.05499}, 2023.

\bibitem{li2022few}
Yiwen Li, Yunguan Fu, Qianye Yang, Zhe Min, et~al.,
\newblock ``Few-shot image segmentation for cross-institution male pelvic organs using registration-assisted prototypical learning,''
\newblock in {\em 2022 IEEE 19th International Symposium on Biomedical Imaging (ISBI)}. IEEE, 2022, pp. 1--5.

\bibitem{hamid_smarttarget}
Sami Hamid, Ian~A Donaldson, Yipeng Hu, Rachael Rodell, Villarini, et~al.,
\newblock ``The smarttarget biopsy trial: a prospective, within-person randomised, blinded trial comparing the accuracy of visual-registration and magnetic resonance imaging/ultrasound image-fusion targeted biopsies for prostate cancer risk stratification,''
\newblock {\em European urology}, vol. 75, no. 5, pp. 733--740, 2019.

\end{thebibliography}

\end{document}